\documentclass[letterpaper]{article} 
\usepackage[submission]{aaai25}  
\usepackage{times}  
\usepackage{helvet}  
\usepackage{courier}  
\usepackage[hyphens]{url}  
\usepackage{graphicx} 
\usepackage{natbib}  
\usepackage{caption} 
\usepackage{algorithm}
\usepackage{algorithmic}

\usepackage{mathbbol}
\usepackage[utf8]{inputenc} 
\usepackage[T1]{fontenc}    
\usepackage{url}            
\usepackage{booktabs}       
\usepackage{amsfonts}       
\usepackage{nicefrac}       
\usepackage{microtype}      
\usepackage{xcolor}         
\usepackage{breqn}
\usepackage{amsmath, amsthm, amssymb}

\usepackage{newfloat}
\usepackage{listings}
\DeclareCaptionStyle{ruled}{labelfont=normalfont,labelsep=colon,strut=off} 
\floatstyle{ruled}
\newfloat{listing}{tb}{lst}{}
\floatname{listing}{Listing}
\title{A practical generalization metric for deep networks benchmarking}
\author{
    Mengqing Huang\textsuperscript{\rm 1}, Hongchuan Yu\textsuperscript{\rm 1}, Jianjun Zhang\textsuperscript{\rm 1}\\
}
\affiliations{
    \textsuperscript{\rm 1}National Centre for Computer Animation,Bournemouth University, UK\\
    mhuang@bournemouth.ac.uk, hyu@bournemouth.ac.uk, jzhang@bournemouth.ac.uk

}

\usepackage{bibentry}

\begin{document}

\maketitle

\begin{abstract}
There is an ongoing and dedicated effort to estimate bounds on the generalization error of deep learning models, coupled with an increasing interest with practical metrics that can be used to experimentally evaluate a model's ability to generalize. This interest is not only driven by practical considerations but is also vital for theoretical research, as theoretical estimations require practical validation. However, there is currently a lack of research on benchmarking the generalization capacity of various deep networks and verifying these theoretical estimations. This paper aims to introduce a practical generalization metric for benchmarking different deep networks and proposes a novel testbed for the verification of theoretical estimations. Our findings indicate that a deep network's generalization capacity in classification tasks is contingent upon both classification accuracy and the diversity of unseen data. The proposed metric system is capable of quantifying the accuracy of deep learning models and the diversity of data, providing an intuitive and quantitative evaluation method — a trade-off point. Furthermore, we compare our practical metric with existing generalization theoretical estimations using our benchmarking testbed. It is discouraging to note that most of the available generalization estimations do not correlate with the practical measurements obtained using our proposed practical metric. On the other hand, this finding is significant as it exposes the shortcomings of theoretical estimations and inspires new exploration.
\end{abstract}

%

\section{Introduction}

Generalization pertains to a model's proficiency in performing well on unseen or new data, focusing on its ability to comprehend and capture underlying data patterns rather than memorizing specific details confined to the training dataset. A well-generalized model showcases excellent performance not solely on the training data but also on novel, previously unseen data.  The assessment of the generalization capability of deep networks has predominantly occurred in supervised learning settings.

Currently, while efforts to establish theoretical bounds for generalization persist, there is an increasing interest in intuitive metrics for experimentally assessing generalization capacity. This trend reflects that many theoretical bounds or capacity measures can be vacuous, inefficient, or even counterproductive in practice. Recent studies have concentrated on interpreting properties associated with deep network generalization, such as robust overfitting in adversarial training (\citet{kim2023NIPS}), exploiting distributional robustness to gauge generalization measures, and combining various complexity measures (\citet{Dziugaite_NIPS20}). Moreover, there is inquiry into whether potential causal relationships between these complexity measures and generalization can be accurately identified (\citet{Jiang2020ICLR}). Additionally, recent advancement in the estimation of non-vacuous generalization bounds (\citet{PAC-Bayes}, \citet{nonvacuousbound}) presented approaches to construct tight generalization bounds, which seek to derive more precise generalization bounds that elucidate the relationship between data fit and model compression. Nonetheless, these theoretical estimations require practical validation as well as a benchmarking framework for practical evaluation and comparison. Moreover, AI faces a reproducibility crisis (\citet{scienceAI}) due to issues such as sharing source codes and data, random number generation, and hyperparameter settings in training. It is essential to provide a public testbed to improve experimental procedures and develop better evaluation methods for benchmarking. This research is not only of theoretical significance but also crucial for addressing practical demands. As of now, there is a lack of relevant research in this area.

This paper introduces a practical metric for measuring generalization capacity (i.e. trade-off point approach) and proposes a novel benchmark testbed for benchmarking various deep networks. Our observations indicate that a deep network's generalization capacity in classical classification scenarios depends on both classification accuracy and the diversity of unseen data. The proposed testbed quantifies model accuracy and test data diversity, providing an intuitive and quantitative assessment.

Moreover, we compare our proposed metric with existing complexity measures using the proposed benchmark testbed. Our findings reveal that most complexity measures do not align with our practical measurements using the proposed practical metric. This discrepancy raises questions about the validity of current theoretical estimations of generalization. The main contributions of this paper include,

\begin{itemize}
  \item Introducing a practical generalization metric for comprehensively benchmarking available deep networks.
  \item Verifying theoretical estimations of generalization through the proposed benchmark testbed. 
\end{itemize}

\section{Related Work}

Our focus lies on the generalization of deep learning models in supervised learning. Current research centers around the estimation of generalization error bounds. There is a growing consensus that traditional approaches in machine learning theory, grounded in worst-case analyses, are inadequate to fully elucidate the generalization of deep learning models (\citet{zhang2021understanding}). This insufficiency is particularly evident when attempting to explain why neural networks demonstrate superior generalization capabilities with over-parametrization (\citet{neyshabur2018towards}). (\citet{ICML23Benjamin}) further introduces a data-dependent fractal dimension to generalisation bound estimations.

A significant work in this direction was done by (\citet{neyshabur2018towards}), which introduced a complexity measure based on unit-wise capacities, resulting in a more precise generalization bound for two-layer ReLU networks. Additionally, (\citet{valle2020generalization}) conducted a comprehensive review of generalization error bound estimation. This review proposed seven desiderata for evaluating generalization in deep learning models and systematically assessed existing approaches for estimating generalization error bounds. These approaches were categorized based on the criteria established by the aforementioned desiderata (\citet{valle2020generalization}).

The first category, data-independent and algorithm-independent, includes algorithms with minimal assumptions and negligible dependence on training data. Notable approaches encompass VC dimension bounds ( \citet{harvey2017nearly}).
The data-dependent and algorithm-independent class involves algorithms with minimal assumptions but reliant on training data, such as the Rademacher complexity bound (\citet{bartlett2002rademacher} and \citet{shawe1997pac}).

Algorithms in the data-independent and algorithm-dependent class carry strong assumptions yet do not depend on the training data, including (\citet{hardt2016train, mou2018generalization, brutzkus2017sgd}).
Finally, the data-dependent and algorithm-dependent category features algorithms with strong assumptions that are dependent on the training data, encompassing methodologies presented in (\citet{barron2019complexity, golowich2018size, neyshabur2017pac, banerjee2020randomized, arora2018optimization, cao2019generalization, zhou2018non, valle2018deep}). Notably, (\citet{Dziugaite17}) introduced the first non-vacuous PAC-Bayes generalization bounds for deep stochastic neural networks on the binary MNIST dataset. Subsequent work by (\citet{PAC-Bayes}, \citet{nonvacuousbound}) proposed new compression approaches for deep networks to construct tighter generalization bounds than have been previously achieved. These endeavors not only hold theoretical significance but also contribute to providing a framework for comprehending deep learning generalization.

Beyond supervised learning, generative models, specifically Generative Adversarial Networks (GANs) (\citet{goodfellow2014generative}), have gained prominence for fitting complex real-world data. A notable observation presented in (\citet{radford2021learning}) revealed that GANs produced synthetic datasets closer to the test set than the training set in the feature space of well-trained deep network classifiers. This finding highlights the potential suitability of GANs for exploring generalization error bound predictions. However, evaluating the generalization capacity of Deep Generative Models poses challenges due to the curse of dimensionality. Moreover, recent studies have highlighted the susceptibility of machine learning models to adversarial attacks (\citet{mustafaICML22}). (\citet{PoursaeedICCV21}) propose Generative Adversarial Training approach to enhance model generalization, robustness against adversarial attacks. Recent studies aim at the adversarial robust leaning (\citet{ACMCS24}). The main concern is that while robust training error can be minimized using various methods, existing algorithms still result in high robust generalization error.

Moreover, the Predicting Generalization in Deep Learning competition (\citet{jiang2020neurips}) held at NeurIPS 2020 featured eight tasks, each with pre-trained deep network classifiers of similar architectures but with differing hyper-parameter settings. This competition applied Conditional Mutual Information to explore the correlation between model complexity and actual generalization gap. While our metric does not compute model complexity, it encompasses dimensions covering various hyperparameter types by introducing robustness and model size, aiming to capture a broad spectrum of hyperparameter variations. In fact, the available complexity measures are not usually consistent with actual generalization gaps in our experiments.
\begin{figure}[ht]
\vskip 0.2in
\begin{center}
\centerline{\includegraphics[width=\columnwidth]{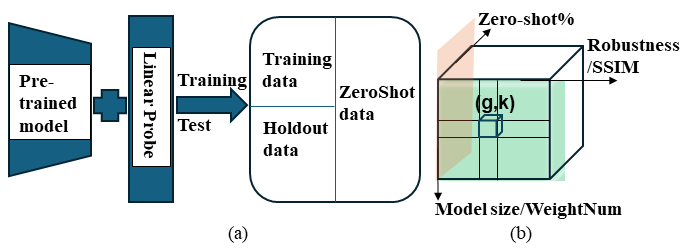}}
\caption{(a)Illustration of Benchmark Testbed; (b)A 3D array consists of cells $(g,k)$, and the pink piece refers to the slice without noise (SSIM=1) and blue piece refers to the slice with zero-shot\%=0.}
\label{Benchmarking bed}
\end{center}
\vskip -0.2in
\end{figure}

\section{Proposed practical generalization metric}
The proposed metric is to measure the generalization capacity of a model through the accuracy (such as classification correct or error rates) and the diversity of test data (such as Kappa) in terms of three factors (i.e. model size, robustness, zero-shot data). Our framework for benchmarking the generalization of deep networks comprises two integral components: the Benchmark Testbed, responsible for producing raw data, and the practical Generalization Metric, which evaluates the model's generalization capacity.

\subsection{Benchmark Testbed}
The proposed benchmark testbed employs the linear probe CLIP structure (\citet{radford2021learning}) to assess how effectively a deep learning model captures essential features within its hidden layers. In our implementation, this involves training a simple linear model, such as logistic regression, on a specific training dataset to fine-tune the tested models. The tested models are always pretrained, and fine-tuned with the linear probe together in our implementation. Notably, since the linear probe cannot capture intricate patterns, high performance indicates that the complexity lies within the features themselves, rather than within the linear probe.
Figure \ref{Benchmarking bed}a illustrates the Benchmark Testbed, where the pretrained model is fine-tuned with a linear probe on specific training data. This fine-tuned model is then evaluated on the holdout data to assess the pretrained model's performance.

Experimentally, the data is divided into two parts: the training data and the holdout data, both sharing the same classes. The pre-trained models are fine-tuned on the training dataset and then tested on the holdout dataset. We gather measured data, specifically ErrorRate and Kappa (defined by Eq.1 and Eq.2), across three distinct dimensions: model size (representing the number of weights), robustness (adding noise and using Structural Similarity Index as a metric, SSIM), and zero-shot capacity (using the percentage of unseen classes).

Notably, the model size dimension can demonstrate the “over-parameterization” effect (\citet{neyshabur2018towards}). Many studies have shown that “over-parameterization” benefits generalization capacity. Although model size does not precisely reflect the architecture of the tested models, it serves as an important indicator for benchmarking purposes.

Regarding the robustness dimension, in deep learning, robustness measures how well a network performs under controlled variations such as noise or distortions, providing insights into the network's ability to generalize effectively (\citet{win2020PGDL}). This concept is extended to adversarial robust learning settings under the umbrella of adversarial robustness. Recent works focus on the generalization gap in robust learning contexts (\citet{Zhang2021ICML,YangNIPS20}). Further exploration of robust generalization challenges in adversarial learning models can be found in (\citet{NIPS22} and \citet{NEURIPS2023kim}). Moreover, (\citet{JACM23}) highlights that "over-parameterization" is also necessary for robust learning. Consequently, robustness is incorporated into our testbed by introducing adversarial samples into the test data.

We use the percentage of unseen classes in the data as the zero-shot dimension to assess zero-shot capacity. It is reasonable that when applying the fine-tuned tested models to the zero-shot data, the percentage of unseen classes in the data serves as an indicator of zero-shot capacity.

This approach results in a three-dimensional array, as shown in Figure \ref{Benchmarking bed}b. Each cell within this array records the distributions of ErrorRate (denoted as “g”) and Kappa metrics (denoted as “k”) across all classes. Different cells within the 3D array correspond to individual settings of the three dimensions. This comprehensive evaluation procedure offers insights into the efficacy of feature extraction within the pre-trained model, allowing an assessment of how well these captured features generalize to new or unseen data.

The generalization gap is defined in (\citet{jiang2020neurips}),
\begin{equation}
\begin{split}
g\left(f_w ; D\right)=\frac{1}{\left|D_{\text {test }}\right|} \sum_{(x, y) \in D_{\text {test }}}  \mathbb{1}\left(f_w(x) \neq y\right) \\
-\frac{1}{\left|D_{\text {train }}\right|} \sum_{(x, y) \in D_{\text {train }}}\mathbb{1}\left(f_w(x) \neq y\right)
\end{split}
\label{accuracy}
\end{equation}
where $w$ denotes a set of model’s weights. Moreover, various hyperparameter types introduce diverse weight values, which results in many variations of some model. Ideally these variations inherit properties of the original model. A rising issue is to capture changes in every single hyperparameter type and measure changes in generalization gap accordingly. In an effort to replicate this random space, (\citet{jiang2020neurips}) selects weight values from a spectrum of hyperparameter types. However, we have another opinion, that is, the variations of some model may be regarded as different models. This is because they may have individual network connection, layers, weights etc. If they are regarded as individuals, our benchmark testbed can test these variations in-depth and streamline model design accordingly.


\subsection{Practical Generalization Metric}
The proposed metric is to seek for a trade-off point to illustrate the generalization of test models as follows.

\textbf{Step 1.} We compute the ErrorRate of individual classes on the test data using Eq.1. It enables the derivation of a distribution of error rates across all classes, while the generalization error typically refers to the overall error rate. We then evaluate the diversity of the test data using the Kappa statistic (\citet{cohen1960coefficient}). In the context of multi-class classification problem, we are dealing with agreement and disagreement among classifier outputs. The Kappa is indeed more robust than simple percentage agreement because it adjusts for the possibility of agreement occurring by chance. This is particularly useful when there is a class imbalance, as chance agreement would be higher for the more frequent classes. Similarly, it also results in a distribution of Kappa across all classes.

Given a dataset with multiple classes, we may divide all the classes into two parts according to the current class $i$, that is, the i-th class and non i-th classes. The classification event is denoted as $h_i(x)=1$ for classifying x into the i-th class or $h_i(x)=-1$ for classifying x not into the i-th class. Similarly, $h_{\bar{i}}(x)=1$ for classifying x into the non i-th classes or $h_{\bar{i}}(x)=-1$ for classifying x not into the non i-th classes. The classification results can be described as $\left\{(x_1,y_1),(x_2,y_2),...,(x_n,y_n)\right\}$, where $y_i\in \left\{-1,1\right\}$ are the class labels of binary classification. The confusion matrix of the $\left\{h_i\right\}$ and $\left\{h_{\Bar{i} }\right\}$ for binary classification is

\begin{center}
\begin{small}
\begin{sc}
\begin{tabular}{|c|l|l|}
\hline & $h_i=1$ & \multicolumn{1}{|c|}{$h_i=-1$} \\
\hline$h_{\bar{i}}=1$ & \#$a$ & \#$c$ \\
\hline$h_{\bar{i}}=-1$ & \#$b$ & \#$d$ \\
\hline
\end{tabular}
\end{sc}
\end{small}
\end{center}
where $\#a$ represents the number of samples predicted as positive in line with the events $h_i$ and $h_{\bar{i}}$, and similarly for $\#b, \#c, \#d$. For example, when the fine-tuned model outputs high but very close probabilities for multiple candidate classes, including the i-th class, this results in conflict. The samples can not be recognized by the model. We thus count them in $"\#a"$. When the fine-tuned model outputs low but very close probabilities for multiple candidate classes, including the i-th class, this results in conflict as well. The samples cannot also be recognized by the model. We thus count them in $"\#d"$. It can be noted that $\#a$ and $\#d$ refer to conflict case numbers while $\#b, \#c$ refer to conflict-free case numbers. It is conceivable that certain samples may go unnoticed by the fine-tuned model due to excessively high loss or low probability in the model outputs. Therefore, we set a threshold to identify such failed samples and count them in “$\#d$”.
The Kappa about the i-th class is defined as,
\begin{equation}
\left\{
\begin{gathered} 
k_i=\frac{p_1-p_2}{1-p_2}\\
p_1=\frac{a+d}{N}, 
p_2=\frac{(a+b)(a+c)+(c+d)(b+d)}{N^2}
\end{gathered} \right.
\label{kappa} 
\end{equation} 
where $N$ denotes the number of total class samples. The average of the Kappas for all the classes may be regarded as the generalization Kappa.

A model with strong generalization capacity should be adaptable to highly diverse data. When the Kappa statistic is high, it indicates that the model is struggling to properly classify samples into different classes, leading to an excessive number of conflict cases. This suggests that the model has low diversity, and consequently, a low generalization capacity. Conversely, if the Kappa statistic is low, it implies that the model exhibits high diversity, and therefore has a high generalization capacity.

\textbf{Step 2.} Within the three dimensions (zero-shot\%, weight number, robustness) of the 3D array, we can calculate two distributions on a cell-wise basis: one related to ErrorRate and the other to Kappa. These calculations are carried out by Eq.\ref{accuracy} for ErrorRate and Eq.\ref{kappa} for Kappa, and are stored within the 3D array (denoted as a pair of "$g$ and $k$" for each cell, see Figure\ref{Benchmarking bed}b).

We depict these two distributions of each cell by three kinds of statistics, i.e., means (denoted as $M$), standard deviations (denoted as $SD$), and $10th$ percentiles (denoted as $^{10}P$). The $10th$ percentile score indicates that 10\% of the trials scored below it. Since smaller means are better in this context, the $10th$ percentiles represent the best performing 10\% of classification outcomes.

We update each cell in the 3D array by these three kinds of statistics with respect to two distributions (i.e., ErrorRate and Kappa) within three dimensions, that is, $M_{g}(ZeroShot, Robust,WeightNum)$, $SD_{g}(ZeroShot, Robust,WeightNum)$, $^{10}P_{g}(ZeroShot, Robust,WeightNum)$ on ErrorRate and  $M_{k}(ZeroShot, Robust,WeightNum)$, $SD_{k}(ZeroShot, Robust,WeightNum)$, $^{10}P_{k}(ZeroShot, Robust,WeightNum) $ on Kappa.


\textbf{Step 3.} We estimate the trade-off point based on the three kinds of statistics within three dimensions in the 3D array. The desired generalization capacity should be achieving high performance of accuracy and diversity by maximizing two dimensions of zero-shot capabilities and robustness, while minimizing the dimension of model size as much as possible.\\
Searching the trade-off point over the 3D array ($3DA$) is described as,
\begin{equation} \label{eq:3}
\begin{aligned}
\underset{(x,y,z) \in 3DA} \min \left( M_g(x, y, z)+SD_g(x, y, z)+{^{10}P_g(x, y, z)} \right.\\ \left. +M_k(x, y, z)+SD_k(x, y, z)+{^{10}P_k(x, y, z)} \right) \\
\text { subject to }\left\{\begin{array}{c}
c_1: x \geqslant ZeroShot_{\min } \\
c_2: y \geqslant Robust _{\min } \\
c_3: z \leqslant WeightNum_{\max}\\
\end{array} \right.
\end{aligned}
\end{equation}
where $(ZeroShot_{\min }, Robust_{\min }, WeightNum_{\max })$ are the given maximum(/minimum) bounds of three dimensions. Particularly, we prefer to maximize (or minimize) these bounds for generalization purpose here. Equation\ref{eq:3} may be converted to a minmax optimization problem as follows,



\begin{equation} \label{eq:4}
\begin{aligned}
\underset{(c_1,c_2,c_3)}{\min} \left\| C \right\|^2 \\
\text{subject to:} \left\{
\begin{array}{l}
\underset{(x,y,z) \in 3DA}{\min} \Big( M_g(x, y, z)+SD_g(x, y, z)+{^{10}P_g(x, y, z)} + \\
M_k(x, y, z)+SD_k(x, y, z)+{^{10}P_k(x, y, z)} \Big) \\
c_1 \geqslant 1-x \\
c_2 \geqslant y \\
c_3 \geqslant z \\
\end{array} 
\right.
\end{aligned}
\end{equation}

where $C=(c_1,c_2,c_3)$ denotes the upper bounds. We apply GEKKO\cite{gekko} to minimize the upper bounds of three dimensions (i.e., ZeroShot, Robust, WeightNum) to approach the trade-off point. Ideally, the resulting $(x,y,z)$ would be equal to the resulting $(c_1,c_2,c_3)$. We always select the resulting $(x,y,z)$ as the trade-off point in practice.


To visualize it, we compute the marginal distributions with respect to three dimensions separately. The marginal distributions with respect to the dimension of $ZeroShot$ is computed as,
\begin{equation}  \label{eq:5}
\left\{\begin{array}{c}
M_g(x \sim 3 D A( ZeroShot ))= \\ 
\sum_{(y, z) \sim 3 D A( Robust,WeightNum )} M_g(x, y, z) \\
SD_g(x \sim 3 D A( ZeroShot ))= \\
\sum_{(y, z) \sim 3 D A( Robust,WeightNum )} SD_g(x, y, z) \\
{^{10}P_g(x \sim 3 D A( ZeroShot ))}= \\
\sum_{(y, z) \sim 3 D A( Robust,WeightNum )} {^{10}P_g(x, y, z)} \\
M_k(x \sim 3  DA ( ZeroShot ))= \\
\sum_{(y, z) \sim 3 D A( Robust,WeightNum )} M_k(x, y, z) \\
SD_k(x \sim 3  DA ( ZeroShot ))= \\
\sum_{(y, z) \sim 3 D A( Robust,WeightNum )} SD_k(x, y, z) \\
{^{10}P_k(x \sim 3  DA ( ZeroShot ))}= \\
\sum_{(y, z) \sim 3 D A( Robust,WeightNum )} {^{10}P_k(x, y, z)}
\end{array}\right.
\end{equation}
There are a total of three sets of marginal distributions separately for three dimensions. Each set illustrates the generalization bounds (referred to as $M_g, SD_g, {^{10}P_g}$) and diversity (referred to as $M_k, SD_k, {^{10}P_k}$) concerning the scale at each dimension specified by the trade-off point, one after another. Theoretical equivalence is expected among these three sets of marginal probabilities at the trade-off point.

In fact, the trade-off point indicates the model’s tolerance on three dimensions at an expected marginal probability level. The area delimited by the trade-off point intuitively and quantitatively illustrates the generalization capacity of the test model.

\section{Benchmarking Tests}
We organise our experiments to illustrate how to determine the Trade-off points by the proposed practical generalisation metric, and then verify the existing complexity measures through the practical measurements based on our testbed. We hope to point out that the proposed benchmark testbed serves solely as an experimental platform to validate existing complexity measures.
\begin{figure*}[ht]
\vspace{-15pt}
\centerline{\includegraphics[width=0.9\textwidth]{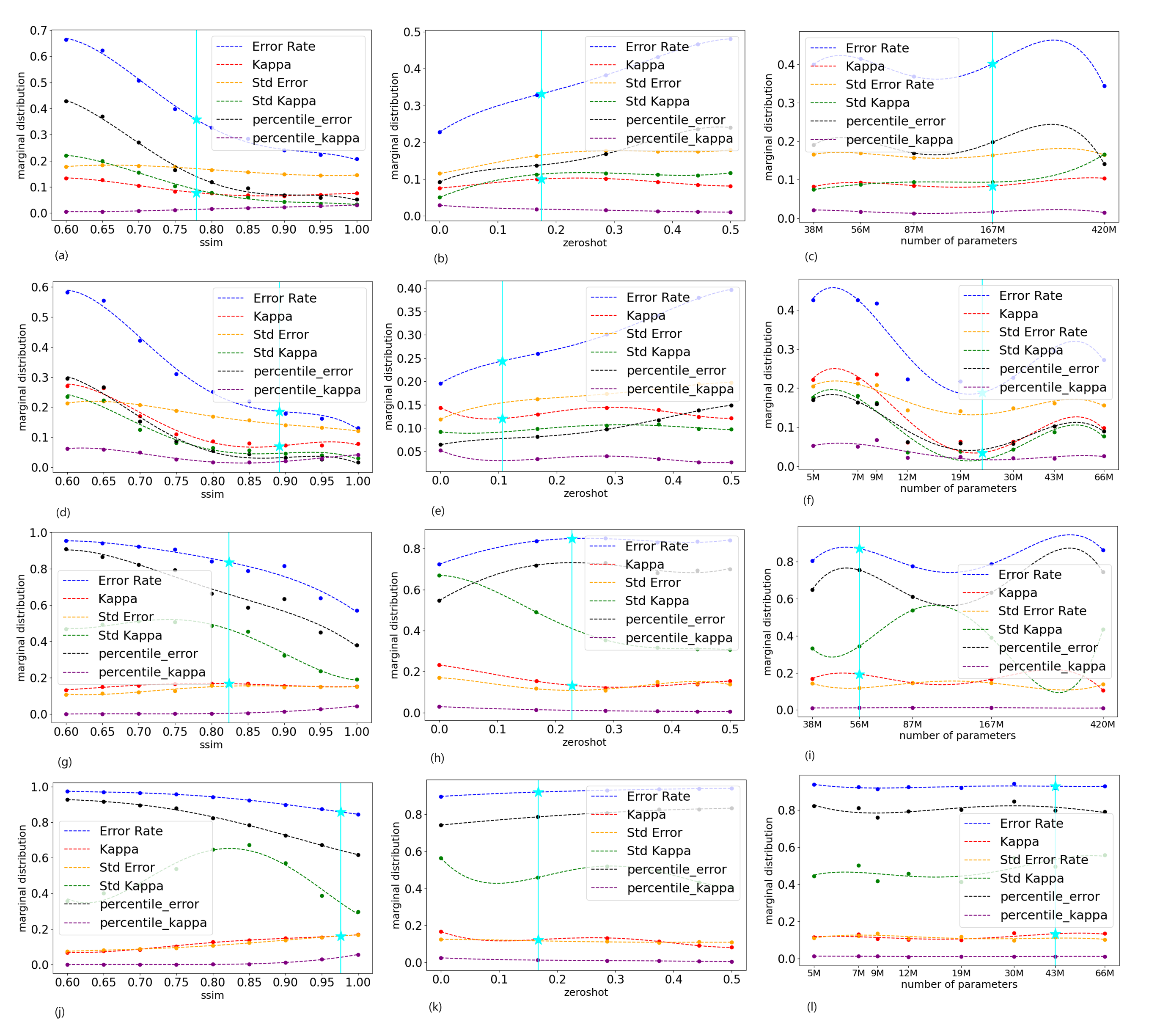}}
\caption{TradeOff points of two kinds models, CLIP and EfficientNet (denoted as $"\star"$). The solid vertical lines indicate the selection of trade-off points on each marginals. (a)-(c) CLIP on ImageNet, (d)-(f) EfficientNet on ImageNet, (g)-(i) CLIP on CIFAR-100, (j)-(l) EfficientNet on CIFAR-100}. 
\label{Tradeoff point.}
\end{figure*}

\subsection{Data and Test Models}
We use CIFAR-100 (\citet{krizhevsky2009learning}) and ImageNet datasets \citep{ILSVRC15} for fine-tuning and tests. In our experiments, we pick up 50 classes for training and the rest 50 classes for the zero-shot scenario tests from CIFAR-100. We randomly select 100 object classes from ImageNet. Similarly, we divide it into two parts, i.e., 50 classes for training and the other 50 classes for tests. These two datasets are widely used in deep learning applications. The primary difference is the image size; ImageNet images are larger than those in CIFAR-100. Larger images in ImageNet provide more data, which generally leads to better learning outcomes. In contrast, the smaller images in CIFAR-100 often result in ambiguity, where additional context is necessary to accurately interpret the images. In addition, we apply augmentation approaches to these datasets to generate unseen data or classes in case that the pretrained models have seen data in their previous training.

We select the CLIP and EfficientNet models for benchmarking tests since they both share similar architecture. They have some connections as well as differences. We use 5 pre-trained CLIP models from \citet{radford2021learning} and 8 EfficientNet models from \citet{tan2019efficientnet}. Table \ref{tab:model_params} shows the pre-trained model sizes of CLIP and EfficientNet respectively. Although these pre-trained models have been optimised, they still need to be fine-tuned with the linear probe on the training data in advance. We only use the weight number of each model as the dimension of model size in the experiments, neglecting the other issues such as layers, depth, the change of structure, so that the pre-trained models line up in an "over-parameterization" way. We hope to have an insight to the generalisation capacity of these two kinds of pre-trained models, i.e. CLIP group and EfficientNet group. Moreover, the test data is added noises for robustness tests. To quantify noise levels, we employ the Autoencoder to the test data to generate noisy data and use the Structural SIMilarity (SSIM) Index metric to control noise levels. When SSIM is decreasing towards zero, the noise level is increasing. All the experiments work on a Workstation with Nvidia 12G RTX2080.
All the data, models, and benchmarking results are available on GitHub (https://...).

\begin{table}[htbp]
\centering

\begin{tabular}{lccc}
\toprule
\textbf{EfficientNet} & \textbf{\# Params}& \textbf{CLIP} & \textbf{\# Params}\\
\midrule
efficientnet-b0 & 5.3M &RN50 & 38M \\
efficientnet-b1 & 7.8M &RN101 & 56M \\
efficientnet-b2 & 9.2M &RN50x4 & 87M \\
efficientnet-b3 & 12M & RN50x16 & 167M \\
efficientnet-b4 & 19M & RN50x64 & 420M \\
efficientnet-b5 & 30M & ViT-B/32 & 87M \\
efficientnet-b6 & 43M & ViT-B/16 & 86M \\
efficientnet-b7 & 66M & ViT-L/14 & 304M \\

\bottomrule

\end{tabular}
\caption{Pretrained Models' Parameters}
\label{tab:model_params}
\end{table}

\subsection{Trade-Off points of CLIP and EfficientNet}

The pre-trained CLIP models (i.e. RNxxx) and EfficientNet models are CNN-based (see Table\ref{tab:model_params}). For comparison, the CLIP ViT-xxx models are not taken into account here. 

\textbf{Step 1. Collect ErrorRate and Kappa data of both kinds of test models}\\
We test the pretrained models of CLIP and EfficientNet on test data across three dimensions (i.e., zero-shot\%, weight number, SSIM) and store the error rates and Kappas for each class in each cell of a 3D array.

\textbf{Step 2. Update 3D Array}\\
We compute three kinds of statistics related to the distributions of ErrorRate and Kappa across all classes, i.e., means, standard derivations, $10th$ percentiles, and update them cell-wise in the 3D array.

\textbf{Step 3. Trade-Off point} \\
We compute the trade-off points by Eq.\ref{eq:4} and visualize the trade-off points by Eq.\ref{eq:5} based on three pairs of marginal distributions, as shown in Figure 3. The trade-off points of CLIP and EfficientNet on CIFAR1-100 and ImageNet respectively are shown in Table 2 and 3.

\begin{table}[t]
\label{Bounds on Imagenet}
\centering
\setlength{\tabcolsep}{0.7mm} 
\begin{tabular}{|c|c|c|}
\hline \text { MODEL TYPE } & \text { CLIP } & \text { EFFICIENT NET } \\
\hline \text { GENERALIZATION BOUND } & 0.364 & 0.206 \\
\hline \text { DIVERSITY BOUND } & 0.087 & 0.075 \\
\hline \text { SSIM(lower bound) } & 0.779 & 0.891 \\
\hline \text { ZEROSHOT(upper bound) } & 0.175 & 0.106 \\
\hline \text { MODEL SIZE(lower bound) } & 167M & 23M \\
\hline
\end{tabular}%
\caption{TradeOff points on ImageNet}
\end{table}
\begin{table}[t]
\label{Bounds on CIFAR100}
\centering
\setlength{\tabcolsep}{0.7mm} 
\begin{tabular}{|c|c|c|}
\hline \text { MODEL TYPE } & \text { CLIP } & \text { EFFICIENT NET } \\
\hline \text { GENERALIZATION BOUND } & 0.852 & 0.902 \\
\hline \text { DIVERSITY BOUND } & 0.164 & 0.139 \\
\hline \text { SSIM(lower bound) } & 0.824 & 0.976 \\
\hline \text { ZEROSHOT(upper bound) } & 0.228 & 0.166 \\
\hline \text { MODEL SIZE(lower bound) } & 56M& 43M\\
\hline
\end{tabular}%
\caption{TradeOff points on CIFAR-100}
\end{table}
It can be noted that,
{\bf (1) CLIP model does not outperform the EfficientNet model}. Comparing the trade-off points in Tables 2 and 3, CLIP's generalization bound exceeds EfficientNet's by up to 0.16 on ImageNet, and its diversity bound is higher by up to 0.01. On CIFAR-100, CLIP's generalization bound is lower by up to 0.05, while its diversity bound is higher by up to 0.02. Although the CLIP's SSIM(lower bound) and ZeroShot(upper bound) are better than EfficientNet's, EfficientNet's model size is much smaller than CLIP's.

Comparing the marginal distributions in Figure 2, the trends of CLIP and EfficientNet (including ErrorRate and Kappa) 
on SSIM and ZeroShot dimensions are similar (see the 1st and 2nd columns in Fig.2). However, the trends for CLIP are opposite to those for EfficientNet on the model size dimension (see the 3rd column). EfficientNet is a compact CNN architecture that uses a compound coefficient to scale models effectively, rather than randomly scaling width, depth, or resolution. Compared to the pretrained CLIP models, EfficientNet models are much smaller and more sensitive to changes in model size. Consequently, the CLIP model does not show an advantage against the EfficientNet model.

\begin{figure*}[!htbp]
\vspace{-15pt}
\centerline{\includegraphics[width=1\textwidth]{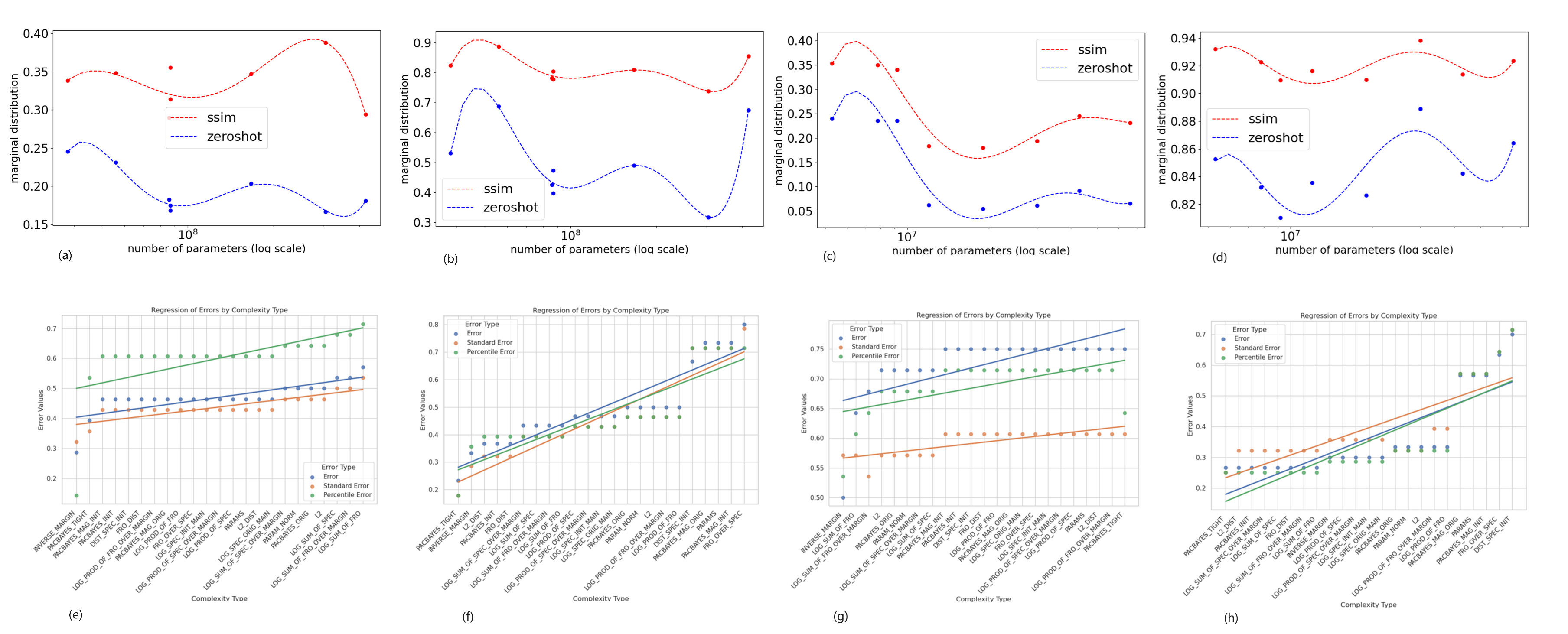}}
\caption{Upper row: Four marginal
probabilities of two slices with respect to the dimension $WeightNum$: (a) CLIP (b) EfficientNet on ImageNet, (c) CLIP (d) EfficientNet on CIFAR-100. Bottom row: Scatter plots of the sign-errors: (e) related to SSIM on ImageNet, (f) related to ZeroShot on ImageNet, (g) related to SSIM on CIFAR-100, (h) related to ZeroShot on CIFAR-100.} 
\label{2margin}
\end{figure*}

A reasonable explanation is that the available pretrained CLIP models include both CNN and Transformer types. Here, we selected CNN-based pretrained CLIP models, but ViT-based CLIP models might perform better.\\
{\bf (2) difference between datasets}. It can be noted that the generalisation and diversity bounds on ImageNet are much less than on CIFAR-100 in Table 2 and 3. Moreover, it can be noted that STD Kappas on CIFAR-100 are obviously more than those on ImageNet in Figure 2. This indicates that the results on ImageNet are always better than on CIFAR-100 since big images can provide more data.

\subsection{Consistency check with existing Generalisation Estimations}

\citet{Dziugaite_NIPS20} and recent work (\citet{nonvacuousbound}) present 23 generalization measures, which we apply to all the pre-trained models listed in Table \ref{tab:model_params}. Our goal is to assess the consistency between existing theoretical estimations and actual measures, and to evaluate agreement/disagreement rates among the available theoretical approaches. For comparison, we focus on two slices of the 3D array rather than the entire array: one for data without robustness and another for data without zero-shot capacity (see the pink and blue sections in Figure \ref{Benchmarking bed}b). This allows us to obtain two distributions of error rates—one for robustness and model size dimensions, and the other for zero-shot and model size dimensions. Note that Kappa is not considered here, as the available complexity estimations focus on generalization error rates. We conduct the consistency check between theoretical estimations and actual measures using these two distributions.

The dimensions of robustness and zero-shot capacity are regarded as two independent factors. We compute two marginal probabilities of these two slices with respect to the dimension of $WeightNum$ (i.e., distributions with respect to $WeightNum$) as below,

\begin{equation} \label{eq:dtr}
\left\{\begin{array}{c}
dtr_g(z \sim 2DSLICE( WeightNum ))=\\ 
\sum_{(y) \sim 2DSLICE( Robust )} dtr_g(y, z) \\
dtr_g(z \sim 2DSLICE( WeightNum ))= \\
\sum_{(x) \sim 2DSLICE( ZeroShot )} dtr_g(x, z)
\end{array}\right.
\end{equation}
Figure\ref{2margin}(a)-(d) shows these marginals based on ImageNet and CIFAR-100 respectively.
Then, we compute the empirical sign-error of generalization in terms of the resulting marginal probabilities Eq.\ref{eq:dtr} as below,
\begin{equation}
\begin{array}{c}
SE_{g} = \frac{1}{2}\mathbb{E}_{(w,w')\sim  \left\{ WeightNum \right\}} 
\left[ 1-sgn(dtr_{g}(w) \right. \\ \left. -dtr_{g}(w'))sgn(C(w)-C(w')) \right] 
\end{array}
\end{equation}
where $w$ and $w'$ denote two different $WeightNum$s from the range of model size; $C(.)$ denotes the complexity measures computed using (\citet{Dziugaite_NIPS20,nonvacuousbound}). If the practical measures ($dtr_g$) and complexity measures ($C$) exhibit consistent changes, the sign-error ($SE_g$) approaches zero. Conversely, inconsistent changes lead to an $SE_g$  approaching one. Consequently, an $SE_g$  exceeding 0.5 indicates a potential mismatch between theoretical estimation and actual measures. Figure\ref{2margin}(e)-(h) visualizes the distributions of sign-errors through scatter plots.

It can be noted that {\bf most of generalisation bound estimations are not consistent with actual measures.}

Regarding the robustness dimension (SSIM), although Figure\ref{2margin}e shows that $30\%$ of $SE_g$  error rates exceed 0.5, Figure\ref{2margin}g indicates that all $SE_g$ values are above 0.5. Furthermore, in both Figure\ref{2margin}e and \ref{2margin}g, the $SE_g$  values for the $10th$ percentile are all greater than 0.5, implying that the top-performing $10\%$ of cases have an error rate exceeding $50\%$. This highlights a significant issue with the reliability of the estimation.
For the ZeroShot dimension, Figure\ref{2margin}f shows that $43\%$ of $SE_g$ error rates exceed 0.5, while Figure\ref{2margin}h indicates that only $21\%$ exceed 0.5. This suggests that the estimation performs better in the ZeroShot dimension compared to robustness. However, most of $SE_g$ of $10th$ percentiles in Figure\ref{2margin}f and \ref{2margin}h are still more than 0.3. The estimations’ reliability is questionable.


\section{Conclusions}

This paper introduces a practical generalization metric for benchmarking diverse deep networks and presents a novel testbed to validate theoretical estimations empirically. By identifying a quantifiable trade-off point, we establish a reliable indicator of model generalization capacity. Our results show a misalignment between existing generalization theories and our practical measurements.

{\bf Limitations}. This paper is limited to CLIP (CNN-based) and EfficientNet models. To enhance benchmarking, a broader range of architectures is required. We have initiated a public GitHub repository for deep network benchmarking and encourage contributions to expand the dataset and foster further theoretical and practical research.

Furthermore, we will organise a comprehensive generalization benchmarking competition for deep networks. This future endeavor seeks to provide developers with a baseline platform to test new theories, thereby enhancing the understanding of why deep neural networks generalize. The benchmarking testbed will facilitate rigorous analyses, enabling developers to assess how well these theories align with the complexities observed in real-world models.

\bigskip

\bibliography{aaai25}

\begin{thebibliography}{42}
\providecommand{\natexlab}[1]{#1}

\bibitem[{Arora, Cohen, and Hazan(2018)}]{arora2018optimization}
Arora, S.; Cohen, N.; and Hazan, E. 2018.
\newblock On the optimization of deep networks: Implicit acceleration by overparameterization.
\newblock In \emph{International Conference on Machine Learning}, 244--253. PMLR.

\bibitem[{Banerjee, Chen, and Zhou(2020)}]{banerjee2020randomized}
Banerjee, A.; Chen, T.; and Zhou, Y. 2020.
\newblock De-randomized pac-bayes margin bounds: Applications to non-convex and non-smooth predictors.
\newblock \emph{arXiv preprint arXiv:2002.09956}.

\bibitem[{Barron and Klusowski(2019)}]{barron2019complexity}
Barron, A.~R.; and Klusowski, J.~M. 2019.
\newblock Complexity, statistical risk, and metric entropy of deep nets using total path variation.
\newblock \emph{arXiv preprint arXiv:1902.00800}.

\bibitem[{Bartlett and Mendelson(2002)}]{bartlett2002rademacher}
Bartlett, P.~L.; and Mendelson, S. 2002.
\newblock Rademacher and Gaussian complexities: Risk bounds and structural results.
\newblock \emph{Journal of Machine Learning Research}, 3(Nov): 463--482.

\bibitem[{Beal et~al.(2018)Beal, Hill, Martin, and Hedengren}]{gekko}
Beal, L.; Hill, D.; Martin, R.; and Hedengren, J. 2018.
\newblock GEKKO Optimization Suite.
\newblock \emph{Processes}, 6(8): 106.

\bibitem[{Brutzkus et~al.(2017)Brutzkus, Globerson, Malach, and Shalev-Shwartz}]{brutzkus2017sgd}
Brutzkus, A.; Globerson, A.; Malach, E.; and Shalev-Shwartz, S. 2017.
\newblock SGD learns over-parameterized networks that provably generalize on linearly separable data.
\newblock \emph{arXiv preprint arXiv:1710.10174}.

\bibitem[{Bubeck and Sellke(2023)}]{JACM23}
Bubeck, S.; and Sellke, M. 2023.
\newblock A Universal Law of Robustness via Isoperimetry.
\newblock \emph{J. ACM}, 70(2).

\bibitem[{Cao and Gu(2019)}]{cao2019generalization}
Cao, Y.; and Gu, Q. 2019.
\newblock Generalization bounds of stochastic gradient descent for wide and deep neural networks.
\newblock \emph{Advances in neural information processing systems}, 32.

\bibitem[{Cohen(1960)}]{cohen1960coefficient}
Cohen, J. 1960.
\newblock A coefficient of agreement for nominal scales.
\newblock \emph{Educational and psychological measurement}, 20(1): 37--46.

\bibitem[{Dupuis, Deligiannidis, and \c{S}im\c{s}ekli(2023)}]{ICML23Benjamin}
Dupuis, B.; Deligiannidis, G.; and \c{S}im\c{s}ekli, U. 2023.
\newblock Generalization bounds using data-dependent fractal dimensions.
\newblock In \emph{Proceedings of the 40th International Conference on Machine Learning}, ICML'23. JMLR.org.

\bibitem[{Dziugaite et~al.(2020)Dziugaite, Drouin, Neal, Rajkumar, Caballero, Wang, Mitliagkas, and Roy}]{Dziugaite_NIPS20}
Dziugaite, G.~K.; Drouin, A.; Neal, B.; Rajkumar, N.; Caballero, E.; Wang, L.; Mitliagkas, I.; and Roy, D.~M. 2020.
\newblock In search of robust measures of generalization.
\newblock In \emph{Proceedings of the 34th International Conference on Neural Information Processing Systems}, NIPS '20. Red Hook, NY, USA: Curran Associates Inc.
\newblock ISBN 9781713829546.

\bibitem[{Dziugaite and Roy(2017)}]{Dziugaite17}
Dziugaite, G.~K.; and Roy, D.~M. 2017.
\newblock Computing Nonvacuous Generalization Bounds for Deep (Stochastic) Neural Networks with Many More Parameters than Training Data.
\newblock In \emph{Proceedings of the 33rd Annual Conference on Uncertainty in Artificial Intelligence (UAI)}.

\bibitem[{Golowich, Rakhlin, and Shamir(2018)}]{golowich2018size}
Golowich, N.; Rakhlin, A.; and Shamir, O. 2018.
\newblock Size-independent sample complexity of neural networks.
\newblock In \emph{Conference On Learning Theory}, 297--299. PMLR.

\bibitem[{Goodfellow et~al.(2014)Goodfellow, Pouget-Abadie, Mirza, Xu, Warde-Farley, Ozair, Courville, and Bengio}]{goodfellow2014generative}
Goodfellow, I.; Pouget-Abadie, J.; Mirza, M.; Xu, B.; Warde-Farley, D.; Ozair, S.; Courville, A.; and Bengio, Y. 2014.
\newblock Generative adversarial nets.
\newblock \emph{Advances in neural information processing systems}, 27.

\bibitem[{Hardt, Recht, and Singer(2016)}]{hardt2016train}
Hardt, M.; Recht, B.; and Singer, Y. 2016.
\newblock Train faster, generalize better: Stability of stochastic gradient descent.
\newblock In \emph{International conference on machine learning}, 1225--1234. PMLR.

\bibitem[{Harvey, Liaw, and Mehrabian(2017)}]{harvey2017nearly}
Harvey, N.; Liaw, C.; and Mehrabian, A. 2017.
\newblock Nearly-tight VC-dimension bounds for piecewise linear neural networks.
\newblock In \emph{Conference on learning theory}, 1064--1068. PMLR.

\bibitem[{Hutson(2018)}]{scienceAI}
Hutson, M. 2018.
\newblock Artificial intelligence faces reproducibility crisis.
\newblock \emph{Science}, 359(6377): 725--726.

\bibitem[{Jiang et~al.(2020{\natexlab{a}})Jiang, Foret, Yak, Roy, Mobahi, Dziugaite, Bengio, Gunasekar, Guyon, and Neyshabur}]{jiang2020neurips}
Jiang, Y.; Foret, P.; Yak, S.; Roy, D.~M.; Mobahi, H.; Dziugaite, G.~K.; Bengio, S.; Gunasekar, S.; Guyon, I.; and Neyshabur, B. 2020{\natexlab{a}}.
\newblock Neurips 2020 competition: Predicting generalization in deep learning.
\newblock \emph{arXiv preprint arXiv:2012.07976}.

\bibitem[{Jiang et~al.(2020{\natexlab{b}})Jiang, Neyshabur, Mobahi, Krishnan, and Bengio}]{Jiang2020ICLR}
Jiang, Y.; Neyshabur, B.; Mobahi, H.; Krishnan, D.; and Bengio, S. 2020{\natexlab{b}}.
\newblock Fantastic Generalization Measures and Where to Find Them.
\newblock In \emph{International Conference on Learning Representations}.

\bibitem[{Kim et~al.(2023{\natexlab{a}})Kim, Park, Choi, and Lee}]{kim2023NIPS}
Kim, H.; Park, J.; Choi, Y.; and Lee, J. 2023{\natexlab{a}}.
\newblock Fantastic Robustness Measures: The Secrets of Robust Generalization.
\newblock In \emph{Thirty-seventh Conference on Neural Information Processing Systems}.

\bibitem[{Kim et~al.(2023{\natexlab{b}})Kim, Park, Choi, and Lee}]{NEURIPS2023kim}
Kim, H.; Park, J.; Choi, Y.; and Lee, J. 2023{\natexlab{b}}.
\newblock Fantastic Robustness Measures: The Secrets of Robust Generalization.
\newblock In Oh, A.; Naumann, T.; Globerson, A.; Saenko, K.; Hardt, M.; and Levine, S., eds., \emph{Advances in Neural Information Processing Systems}, volume~36, 48793--48818. Curran Associates, Inc.

\bibitem[{Krizhevsky, Hinton et~al.(2009)}]{krizhevsky2009learning}
Krizhevsky, A.; Hinton, G.; et~al. 2009.
\newblock Learning multiple layers of features from tiny images.

\bibitem[{Li et~al.(2022)Li, Jin, Zhong, Hopcroft, and Wang}]{NIPS22}
Li, B.; Jin, J.; Zhong, H.; Hopcroft, J.; and Wang, L. 2022.
\newblock Why robust generalization in deep learning is difficult: Perspective of expressive power.
\newblock \emph{Advances in Neural Information Processing Systems}, 35: 4370--4384.

\bibitem[{Lotfi et~al.(2024)Lotfi, Finzi, Kapoor, Potapczynski, Goldblum, and Wilson}]{PAC-Bayes}
Lotfi, S.; Finzi, M.; Kapoor, S.; Potapczynski, A.; Goldblum, M.; and Wilson, A.~G. 2024.
\newblock PAC-Bayes compression bounds so tight that they can explain generalization.
\newblock In \emph{Proceedings of the 36th International Conference on Neural Information Processing Systems}, NIPS '22. Red Hook, NY, USA: Curran Associates Inc.
\newblock ISBN 9781713871088.

\bibitem[{Mou et~al.(2018)Mou, Wang, Zhai, and Zheng}]{mou2018generalization}
Mou, W.; Wang, L.; Zhai, X.; and Zheng, K. 2018.
\newblock Generalization bounds of sgld for non-convex learning: Two theoretical viewpoints.
\newblock In \emph{Conference on Learning Theory}, 605--638. PMLR.

\bibitem[{Mustafa, Lei, and Kloft(2022)}]{mustafaICML22}
Mustafa, W.; Lei, Y.; and Kloft, M. 2022.
\newblock On the Generalization Analysis of Adversarial Learning.
\newblock In Chaudhuri, K.; Jegelka, S.; Song, L.; Szepesvari, C.; Niu, G.; and Sabato, S., eds., \emph{Proceedings of the 39th International Conference on Machine Learning}, volume 162 of \emph{Proceedings of Machine Learning Research}, 16174--16196. PMLR.

\bibitem[{Natekar and Sharma(2020)}]{win2020PGDL}
Natekar, P.; and Sharma, M. 2020.
\newblock Representation Based Complexity Measures for Predicting Generalization in Deep Learning.
\newblock arXiv:2012.02775.

\bibitem[{Neyshabur, Bhojanapalli, and Srebro(2017)}]{neyshabur2017pac}
Neyshabur, B.; Bhojanapalli, S.; and Srebro, N. 2017.
\newblock A pac-bayesian approach to spectrally-normalized margin bounds for neural networks.
\newblock \emph{arXiv preprint arXiv:1707.09564}.

\bibitem[{Neyshabur et~al.(2018)Neyshabur, Li, Bhojanapalli, LeCun, and Srebro}]{neyshabur2018towards}
Neyshabur, B.; Li, Z.; Bhojanapalli, S.; LeCun, Y.; and Srebro, N. 2018.
\newblock Towards understanding the role of over-parametrization in generalization of neural networks.
\newblock \emph{arXiv preprint arXiv:1805.12076}.

\bibitem[{Poursaeed et~al.(2021)Poursaeed, Jiang, Yang, Belongie, and Lim}]{PoursaeedICCV21}
Poursaeed, O.; Jiang, T.; Yang, H.; Belongie, S.; and Lim, S. 2021.
\newblock Robustness and Generalization via Generative Adversarial Training.
\newblock In \emph{2021 IEEE/CVF International Conference on Computer Vision (ICCV)}, 15691--15700. Los Alamitos, CA, USA: IEEE Computer Society.

\bibitem[{Radford et~al.(2021)Radford, Kim, Hallacy, Ramesh, Goh, Agarwal, Sastry, Askell, Mishkin, Clark et~al.}]{radford2021learning}
Radford, A.; Kim, J.~W.; Hallacy, C.; Ramesh, A.; Goh, G.; Agarwal, S.; Sastry, G.; Askell, A.; Mishkin, P.; Clark, J.; et~al. 2021.
\newblock Learning transferable visual models from natural language supervision.
\newblock In \emph{International conference on machine learning}, 8748--8763. PMLR.

\bibitem[{Russakovsky et~al.(2015)Russakovsky, Deng, Su, Krause, Satheesh, Ma, Huang, Karpathy, Khosla, Bernstein, Berg, and Fei-Fei}]{ILSVRC15}
Russakovsky, O.; Deng, J.; Su, H.; Krause, J.; Satheesh, S.; Ma, S.; Huang, Z.; Karpathy, A.; Khosla, A.; Bernstein, M.; Berg, A.~C.; and Fei-Fei, L. 2015.
\newblock {ImageNet Large Scale Visual Recognition Challenge}.
\newblock \emph{International Journal of Computer Vision (IJCV)}, 115(3): 211--252.

\bibitem[{Sanae~Lotfi(2023)}]{nonvacuousbound}
Sanae~Lotfi, Y. K. T. R. M. G. A.~W., Marc~Finzi. 2023.
\newblock Non-Vacuous Generalization Bounds for Large Language Models.
\newblock In \emph{Proceedings of Workshop Mathematics of Modern Machine Learning (M3L) of the 36th International Conference on Neural Information Processing Systems}, Workshop of NIPS '23. Red Hook, NY, USA: Curran Associates Inc.

\bibitem[{Shawe-Taylor and Williamson(1997)}]{shawe1997pac}
Shawe-Taylor, J.; and Williamson, R.~C. 1997.
\newblock A PAC analysis of a Bayesian estimator.
\newblock In \emph{Proceedings of the tenth annual conference on Computational learning theory}, 2--9.

\bibitem[{Tan and Le(2019)}]{tan2019efficientnet}
Tan, M.; and Le, Q. 2019.
\newblock Efficientnet: Rethinking model scaling for convolutional neural networks.
\newblock In \emph{International conference on machine learning}, 6105--6114. PMLR.

\bibitem[{Valle-Perez, Camargo, and Louis(2018)}]{valle2018deep}
Valle-Perez, G.; Camargo, C.~Q.; and Louis, A.~A. 2018.
\newblock Deep learning generalizes because the parameter-function map is biased towards simple functions.
\newblock \emph{arXiv preprint arXiv:1805.08522}.

\bibitem[{Valle-P{\'e}rez and Louis(2020)}]{valle2020generalization}
Valle-P{\'e}rez, G.; and Louis, A.~A. 2020.
\newblock Generalization bounds for deep learning.
\newblock \emph{arXiv preprint arXiv:2012.04115}.

\bibitem[{Xiong et~al.(2024)Xiong, Tegegn, Sarin, Pal, and Rubin}]{ACMCS24}
Xiong, P.; Tegegn, M.; Sarin, J.~S.; Pal, S.; and Rubin, J. 2024.
\newblock It Is All about Data: A Survey on the Effects of Data on Adversarial Robustness.
\newblock \emph{ACM Comput. Surv.}, 56(7).

\bibitem[{Yang et~al.(2020)Yang, Rashtchian, Zhang, Salakhutdinov, and Chaudhuri}]{YangNIPS20}
Yang, Y.-Y.; Rashtchian, C.; Zhang, H.; Salakhutdinov, R.; and Chaudhuri, K. 2020.
\newblock A closer look at accuracy vs. robustness.
\newblock In \emph{Proceedings of the 34th International Conference on Neural Information Processing Systems}, NIPS '20. Red Hook, NY, USA: Curran Associates Inc.
\newblock ISBN 9781713829546.

\bibitem[{Zhang et~al.(2021{\natexlab{a}})Zhang, Cai, Lu, He, and Wang}]{Zhang2021ICML}
Zhang, B.; Cai, T.; Lu, Z.; He, D.; and Wang, L. 2021{\natexlab{a}}.
\newblock Towards Certifying L-infinity Robustness using Neural Networks with L-inf-dist Neurons.
\newblock In \emph{International Conference on Machine Learning}.

\bibitem[{Zhang et~al.(2021{\natexlab{b}})Zhang, Bengio, Hardt, Recht, and Vinyals}]{zhang2021understanding}
Zhang, C.; Bengio, S.; Hardt, M.; Recht, B.; and Vinyals, O. 2021{\natexlab{b}}.
\newblock Understanding deep learning (still) requires rethinking generalization.
\newblock \emph{Communications of the ACM}, 64(3): 107--115.

\bibitem[{Zhou et~al.(2018)Zhou, Veitch, Austern, Adams, and Orbanz}]{zhou2018non}
Zhou, W.; Veitch, V.; Austern, M.; Adams, R.~P.; and Orbanz, P. 2018.
\newblock Non-vacuous generalization bounds at the imagenet scale: a PAC-bayesian compression approach.
\newblock \emph{arXiv preprint arXiv:1804.05862}.

\end{thebibliography}



\end{document}